\documentclass[preprint,12pt]{elsarticle}

\usepackage[normalem]{ulem}
\useunder{\uline}{\ul}{}
\usepackage{amssymb}
\usepackage{amsmath}
\usepackage{float}
\usepackage{hyperref} 

\journal{Medical Image Analysis}

\begin{document}

\begin{frontmatter}



\title{SPADE: Spatial Transcriptomics and Pathology Alignment Using a Mixture of Data Experts for an Expressive Latent Space}




\author[label1,label3,label4]{Ekaterina Redekop\fnref{equal}} 
\author[label1,label3,label5]{Mara Pleasure\fnref{equal}}
\fntext[equal]{These authors contributed equally to this work.}
\author[label1,label3,label4]{Zichen Wang}
\author[label1,label3,label5]{Vedrana Ivezic}
\author[label2]{Kimberly E. Flores}
\author[label2]{Benjamin Emert}
\author[label2]{Anthony Sisk}
\author[label1,label3]{William Speier}
\author[label1,label2,label3,label6]{Corey W. Arnold\corref{cor1}}
\cortext[cor1]{Corresponding author: \href{mailto:cwarnold@mednet.ucla.edu}{cwarnold@mednet.ucla.edu}}
\affiliation[label1]{organization={Biomedical AI Research Lab, University of California, Los Angeles},
            addressline={24 Westwood Blvd}, 
            city={Los Angeles},
            postcode={90024}, 
            state={CA},
            country={United States}}

\affiliation[label2]{organization={Department of Pathology, University of California, Los Angeles},
            city={Los Angeles},
            state={CA},
            country={United States}}
            
\affiliation[label3]{organization={Department of Radiology, University of California, Los Angeles}, city={Los Angeles},
            state={CA},
            country={United States}}

\affiliation[label4]{organization={Department of Bioengineering, University of California, Los Angeles}, city={Los Angeles},
            state={CA},
            country={United States}}

\affiliation[label5]{organization={UCLA Medical Informatics Home Area, University of California, Los Angeles}, city={Los Angeles},
            state={CA},
            country={United States}}

\affiliation[label6]{organization={Department of Computational Medicine, University of California, Los Angeles}, city={Los Angeles},
            state={CA},
            country={United States}}            
\begin{abstract}
The rapid growth of digital pathology and advances in self-supervised deep learning have enabled the development of foundational models for various pathology tasks across diverse diseases. While multimodal approaches integrating diverse data sources have emerged, a critical gap remains in the comprehensive integration of whole-slide images (WSIs) with spatial transcriptomics (ST), which is crucial for capturing critical molecular heterogeneity beyond standard hematoxylin \& eosin (H\&E) staining. We introduce SPADE, a foundation model that integrates histopathology with ST data to guide image representation learning within a unified framework, in effect creating an ST-informed latent space. SPADE leverages a mixture-of-data experts technique, where experts are created via two-stage imaging feature-space clustering using contrastive learning to learn representations of co-registered WSI patches and gene expression profiles. Pre-trained on the comprehensive HEST-1k dataset, SPADE is evaluated on 20 downstream tasks, demonstrating significantly superior few-shot performance compared to baseline models, highlighting the benefits of integrating morphological and molecular information into one latent space. Code and pretrained weights are available at \href{https://github.com/uclabair/SPADE}{https://github.com/uclabair/SPADE}.
\end{abstract}






\end{frontmatter}

\section{Introduction}
\label{sec:intro}


High-resolution whole slide images (WSIs) have propelled the development of powerful deep learning foundation models in computational pathology, demonstrating robust performance across diverse tissue types and tasks \cite{wang2022cell, li2021multi, saednia2022quantitative, howard2023integration}. These models are typically trained using self-supervision, enabling learning from large unlabeled datasets and producing embeddings robust to institutional variations, including differences in staining procedures and other image-quality factors \cite{wang2024pathology, chen2024towards, chen2022scaling, xu2024whole}.


By visually capturing cellular arrangement, WSIs enable the study of spatial organization and disorganization of cells in tissues, characterizations that are especially relevant in cancer research \cite{moffitt2022emerging, moses2022museum}. In clinical settings, WSIs are commonly stained with hematoxylin \& eosin (H\&E), a two-color stain that highlights nuclei and cytoplasm but offers a limited view of molecular-level heterogeneity \cite{lee2023promise}. As tumor tissues are known to exhibit high variability within and across patients, deciphering the heterogeneity at the molecular level is critical for improving deep learning applications that can more precisely inform diagnosis, treatment, and prognosis \cite{rao2021exploring, xun2023reconstruction}. While H\&E provides crucial morphological insights, its inability to capture molecular heterogeneity limits its utility in fully characterizing tissue complexity.

Spatial transcriptomics addresses this gap by providing spatially resolved gene expression data, allowing for additional molecular context for a given tissue specimen. Although both ST and H\&E data have independently proven useful in various applications, their combined potential for creating a more comprehensive representation learning framework remains unexplored. To this end, we introduce SPADE, a vision-ST foundation model that uses a mixture of experts, each trained via contrastive learning, to unify ST data and H\&E images to produce slide representations that encompass both modalities. SPADE is trained using a large collection of Visium data from the HEST-1K dataset to generate a rich, expressive latent representation space that spans multiple tissue types \cite{jaume2024hest}. We hypothesize that training a contrastive learning model on paired ST and WSI data will create a shared latent space that captures molecularly informed features of tissue organization. This enriched representation is expected to improve performance across various downstream tasks compared to conventional WSI self-supervised learning.

SPADE's architecture, based on a mixture-of-experts, is specifically designed to handle the complexities of spatially resolved paired ST and H\&E data. Each expert is trained on a distinct subset of ST-WSI pairs, identified through a two-step K-means clustering process. First, patches are grouped into K subclusters within each organ, yielding per-organ centroids. Then, these centroids are clustered again across all organs to arrive at a final set of K representative clusters spanning the 515 Visium WSIs. This process generates hard negatives, enhancing training for individual experts and ensuring robust representation learning across multiple tissue types. Our contributions are summarized as follows:\noindent
\begin{enumerate}
\setlength{\itemsep}{0pt}
\item \textbf{Multimodal Representation Learning:}
We propose the first histology foundation model that pretrains on paired H\&E and ST data, producing an enriched latent space for pathology tasks.
\item \textbf{Mixture of Data Experts with Hard Negative Mining:} By clustering WSIs in a two-step K-means process, SPADE identifies challenging examples that enhance discrimination and prevent trivial solutions, especially across diverse organ types.
\item \textbf{Comprehensive Evaluation across 20 Tasks:} We demonstrate SPADE's effectiveness on eight cancer classification tasks, six survival prediction tasks, and six biomarker prediction tasks, showcasing its robust performance in modeling the complexity of tumor samples.
\end{enumerate}

\section{Related work}
\label{sec:related_work}

\subsection{Contrastive learning pretraining for multimodal datasets}
Contrastive learning is a core self-supervised approach for large imaging datasets, which uses the InfoNCE loss, introduced by van den Oord et al. \cite{oord2018representation}, to pull positive pairs of data points closer together while pushing negative pairs apart. Contrastive language-image pretraining (CLIP) extends this technique to multimodal data by using paired images and captions \cite{radford2021learning}. Rather than focusing solely on images, CLIP creates a multimodal embedding space using a symmetric InfoNCE loss that matches paired images and captions as positives and mixes random images and captions as negatives. This symmetric process (image-to-caption and caption-to-image) allows CLIP to develop a latent space where images and their corresponding captions cluster together. 

BLEEP \cite{xie2024spatially} applies the CLIP framework to spatial transcriptomics for gene expression prediction, creating a multimodal model specifically for Visium data but trained on a single tissue type, limiting its generalizability. Another model, TANGLE, recently developed by Jaume et al. \cite{jaume2024transcriptomics} applies the CLIP framework to paired H\&E WSIs and bulk RNA-seq data from the Cancer Genome Atlas (TCGA); however, bulk RNA sequencing requires tissue dissociation and lacks one-to-one matching of gene expression to tissue. SPADE addresses these limitations by training on multiple tissue types using a Mixture of Data Experts approach, utilizing the one-to-one correspondence between 20$\times$ H\&E patches and their spatially resolved ST gene expression profiles to learn rich, molecularly informed patch-level embeddings for multiple tissue types.


\subsection{Clustering for non-trivial contrastive learning}
\label{sec:mode}
SPADE is trained on a dataset that spans multiple organs, leading to a risk of the original CLIP framework finding trivial solutions by contrasting different tissue types with each other instead of learning a meaningful latent space. Because CLIP samples negative pairings from within each batch, highly varied images in a batch can make these pairings easier for the model to learn. Ma et al. proposed a Mixture of Data Experts (MoDE) approach to address the limitations of CLIP using non-trivial negative pairings to train by incorporating a two-step K-means clustering process into training \cite{ma2024mode}. MoDE first clusters semantically similar text captions, then trains a distinct CLIP model within each cluster. By sampling hard negatives from within these similar groups, the model is forced to focus on distinguishing features that differentiate similar captions and their paired images \cite{ma2024mode}. 

Using a two-step K-means procedure, we adapt this hard-negative sampling procedure to Visium data and cluster in the image feature space. In \cite{vo2024automatic} the authors demonstrate how hierarchical clustering can be leveraged to create high-quality representations by progressively refining clusters at different levels of abstraction. Inspired by this, our two-step K-means procedure first clusters data within each organ type, and then clusters the resulting organ-specific centroids. This clustering approach allows our model to learn a more detailed and expressive latent space for each expert while also allowing flexibility to work across cancer types and vastly different tissue makeups.

\section{Methods}
\label{sec:method}
\begin{figure}[H]
    \centering
    
    \includegraphics[width=\textwidth]{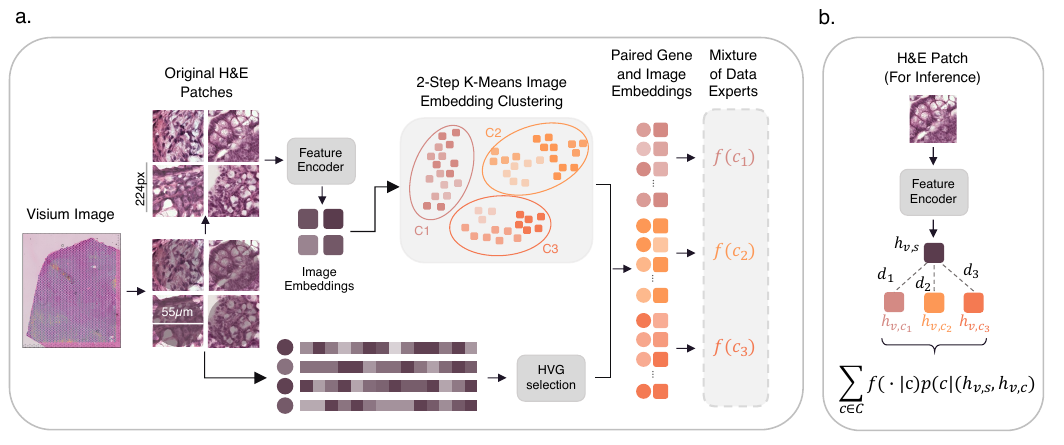} %
    \caption{\textbf{Overview of SPADE workflow}. \textbf{a.} A WSI is segmented and patched into a set of non-overlapping patches. A compressed feature for each patch is obtained through a pre-trained feature encoder. A corresponding gene expression vector is obtained after preprocessing. 2-step K-Means clustering is performed to create data experts across all WSIs in the dataset.  \textbf{b.} For inference, three data experts' routing techniques are evaluated (Section.\ref{sec:slide_repr}).}
    \label{fig:pipeline} 
\end{figure}

We present the \textbf{S}patial transcriptomics and \textbf{P}athology \textbf{A}lignment using a mixture of \textbf{D}ata \textbf{E}xperts (SPADE) framework. (Figure \ref{fig:pipeline}). 
The proposed framework acts as an image encoder, where the extracted embeddings are subsequently aggregated to generate a comprehensive slide-level representation for downstream applications. SPADE is a mixture of data experts $\{f(\cdot|c)\}$, each pretrained using a contrastive objective to align an H\&E patch embedding with a gene expression embedding (Section \ref{sec:bleep}) derived from a specific subset of data $c$ as defined in Section \ref{sec:clustering}. Following the training of data experts (Section \ref{sec:de}), SPADE is comprehensively evaluated on diverse downstream tasks to validate its effectiveness (Section \ref{sec:results}). Implementation details can be found in \ref{sec:expr_details_sup}.

\subsection{Multimodal alignment}
\label{sec:bleep}
SPADE utilizes a contrastive learning framework inspired by CLIP, designed with a modified loss function to learn a bimodal embedding that aligns image patches with their corresponding gene expression data \cite{xie2024spatially}. In contrast to natural images, it is common to sample multiple spots with similar gene expression profiles or image morphology within the same batch, which the contrastive objective should not pull apart. To take into account similar pairs, the similarity-adjusted target $t$ is computed as follows:

\begin{equation}
    t = \frac{\sigma(\text{sim}(h_{x}, h_{x}) + \text{sim}(h_{v}, h_{v}))}{2} \cdot \tau
\end{equation} where $h_{v}$ is an $m$-dimensional
image embedding, $h_{x}$ is an $m$-dimensional
gene expression embedding, and $\tau$ is a temperature parameter. The contrastive objective $\mathcal{L}$ is then updated to align $h_{v}$ and $h_{x}$ in the latent space and is computed with cross-entropy (CE) as follows:
\vspace{-2pt}
\begin{equation}
    \mathcal{L} = \text{mean}\left(\text{CE}\left(\text{sim}(h_v, h_x),t\right) + \text{CE}\left(\text{sim}(h_v, h_x)^{\top}, t^{\top}\right)\right)
\end{equation}


\subsection{Clustering}
\label{sec:clustering}

\begin{figure*}[h!]
    \centering
    
     \includegraphics[width=\textwidth]{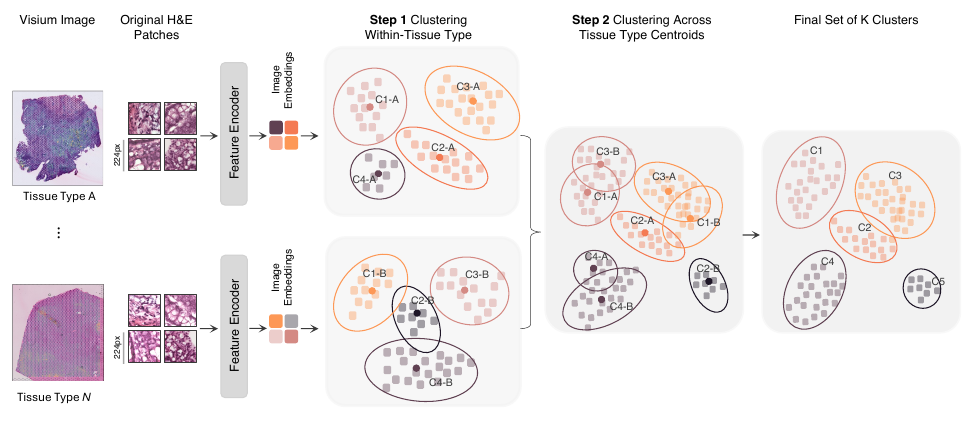} %
    \caption{\textbf{Two-step clustering framework for data expert construction.} 
Each WSI is divided into patches and encoded into patch-level features. 
\textbf{Step 1:} Within each tissue type, K-Means clustering is applied to obtain tissue-specific centroids. 
\textbf{Step 2:} The centroids across tissue types are clustered again to form a final unified set of $K$ clusters (data experts) used for data experts training.}

    \label{fig:2step_clustering} 
\end{figure*}

As shown in Figure~\ref{fig:2step_clustering}, each WSI $X_j, j=1, \dots, J$  from the pretraining cohort $\mathcal{D}$ of size $J$ is first divided into a set of non-overlapping patches \( X_j = \{ x_j^1, \dots, x_j^{N_j} \} \), where each patch \( x_j^n \in \mathbb{R}^{W \times H \times 3} \). A pre-trained foundation model (UNI \cite{chen2024towards} ) serves as the feature encoder \( f_{\text{enc}}(\cdot) \) used to 
extract a compressed embedding from each patch.  This results in a set of embeddings 
\( H_{v,j} = \{ h_{v,j}^1, \dots, h_{v,j}^{N_j} \} \) where \( h_{v,j}^n = f_{\text{enc}}(x_j^n) \in \mathbb{R}^m \).

The goal is to extract a set of image clusters $\{h_{v,1}, ..., h_{v,C}\}$ with $h_{v,c} \in \mathbb{R}^m $ based on the set of patch-level embeddings to formulate conditions $C$ for data expert training. To train these experts, we utilize a two-step K-means clustering approach (see Figure~\ref{fig:2step_clustering}). In the first step, we apply fine-grained clustering utilizing patch-based embeddings within each organ type to identify clusters where the samples share similar semantics. In the second step, we aggregate the fine-grained clusters based on each cluster's centroid to establish coarse-grained clustering, facilitating specialization among data experts across different organs.

Given the large size of the pretraining dataset $\mathcal{D}$, K-means is performed on a subset of the data, as training K-means on the entire dataset may be computationally inefficient. We draw a uniform sample from each organ-specific subset (32 organs in total): $\mathcal{D}^{`}_{i} \sim \mathcal{D}_{i}$ and $|\mathcal{D}^{`}_{i}| \ll |\mathcal{D}_{i}|$, where $1 \leq i \leq 32$. Then fine-grained K-means is performed over each $\mathcal{D}_{i}$: 

\begin{equation}
S_{i} \leftarrow \text{K-means} (\mathcal{D}^{`}_{i}) 
\end{equation}, 
where $S =  \bigcup S_{i}$ - set of learned cluster centers.

A second round of clustering is performed on top of previously found cluster centers $S$:

\begin{equation}
C \leftarrow \text{K-means} (S) 
\end{equation} 

, where each $c \in C$ becomes a condition for a data expert $f(\cdot|c)$.

To determine the optimal number of clusters for both steps, we performed a within-cluster sum of squares (WCSS) analysis and used the elbow method to identify the point where adding more clusters provided diminishing returns. This ensured that our clustering approach captured meaningful biological variation while avoiding excessive fragmentation of the data. More details in \ref{sec:app_clustering}.

\subsection{Data Experts Training}
\label{sec:de}
In this work, we propose to train a set of contrastive models $\{f(\cdot|c)\}$ independently given clusters $C$, such that $c \in C$. The training data for a data expert $\{f(\cdot|c)\}$ is :
\begin{equation}
\mathcal{D}_{c} = \bigcup_{s \in S_{c}} \mathcal{D}_{s}
\end{equation}
, where $\mathcal{D}_{s} = \{d | s = \text {argmin}_{s \in S} (||h_{v,s} - h_{v,c})||^{2}_{2}, d \in \mathcal{D}\}$ is the data assigned for each fine-grained cluster with $h_{v,s}$ and $h_{v,c}$ being the image embeddings for training example
$s$ and fine-grained cluster center $c$ respectively.
$S_{c}$ is a set of fine-grained clusters assigned to each data expert $f(\cdot|c)$, further trained only on $D_{c}$.

\subsection{Slide-level Representation}
\label{sec:slide_repr}
The proposed expert-based approach provides multiple models to select from during inference rather than relying solely on a single contrastive model. We evaluate three distinct weighting techniques for routing data experts by formulating their outputs as a weighted sum for simplicity:
\vspace{-2pt}
\begin{equation}
\sum_{c \in C} f(\cdot|c) \cdot p_{i}(c|(h_{v,s},h_{v,c}))
\end{equation} 
, where $p_{i}(c|(h_{v,s}h_{v,c}))$ represents the normalized weight assigned to each expert, satisfying $\sum_{c \in C} p_{i}(c|(h_{v,s},h_{v,c})) = 1$. Here, $ i \in {1,2,3}$ denotes three weighting techniques being compared, $h_{v,s}$ is the sample's image embedding, and $h_{v,c}$ is the expert`s centroid.
In the first approach, each data sample is routed to a single expert based on the proximity of its embedding to the expert centroids. Specifically:
\begin{equation}
    p_{1}(c | (h_{v,s},h_{v,c})) = 
\begin{cases} 
1 & \text{if } c = \arg\min_{c' \in C} \, d_{c'} \\
0 & \text{otherwise}
\end{cases}
\end{equation} 
where $d_{c'} = (||h_{v,s} - h_{v,c'})||^{2}_{2}$.
In the second approach, the outputs from all experts are averaged for each data sample:
\begin{equation}
    p_{2}(c|(h_{v,s},h_{v,c}))=\frac{1}{C}
\end{equation}

In the third approach, the embedding is compared to all centroids to compute distances $\{d_c'\}_{c' \in C}$, and the outputs are combined using a softmax-based weighting:
\begin{equation}
    p_{3}(c|(h_{v,s},h_{v,c})) 
    = \frac{\exp(-d_{c'})}{\sum_{i \in C} \exp(-d_{i})}.
\end{equation}

After features are obtained for each patch through one of the weighting techniques, a slide-level feature vector can be derived through a learnable aggregation method. Specifically, we employed Attention-Based Multiple Instance Learning (ABMIL) \cite{ilse2018attention}, which allows for aggregating patch-level embeddings into a slide-level representation. In this setup, each patch embedding is treated as an instance, and the model learns to assign attention weights to each instance based on its relevance to the slide-level prediction task. ABMIL leverages an attention mechanism to weight each patch’s contribution, capturing spatial heterogeneity across the slide.
The resulting slide-level feature vector can then be utilized for various downstream tasks, as described in Section \ref{sec:experiemnts}.

\section{Experiments and Results}
\subsection{Datasets}
\label{sec:experiemnts}

Each dataset was processed in a similar manner with a fixed stain normalization and 224$\times$224 patching at magnification 20x. The Visium gene expression data was processed by normalizing each spot to its total count, followed by log normalization. Subsequently, highly variable genes were identified using the Scanpy package \citep{wolf2018scanpy}. The union of the top 50 most highly variable genes from each sample was used for training and prediction, resulting in 7,986 genes following the method outlined in \cite{xie2024spatially}. Additional
details can be found in \textbf{Appendix}.

\textbf{Pretraining}
For pretraining, we utilized the HEST-1k dataset, a diverse collection of 1,229 spatial transcriptomic profiles, each paired with a WSI. HEST-1k spans 153 cohorts, covering 26 organs, and includes 367 cancer samples across 25 types. For this work, we utilized only the Homo sapiens samples obtained using Visium technology, leading to a total of 515 profiles.

\textbf{Subtyping}
We evaluated the proposed approach on six different subtyping tasks: Non-Small Cell Lung Carcinoma (NSCLC) subtyping on
PLCO \cite{zhu2013prostate} and CPTAC \cite{grossman2016toward} (two classes), ISUP grading based on Prostate cancer grade assessment (PANDA) challenge \cite{bulten2022artificial} (six
classes), classification of breast cancer metastases in lymph nodes (Camelyon16) \cite{bejnordi2017diagnostic, litjens20181399} (three classes), ovarian cancer subtypes classification (UBC-OCEAN) challenge \cite{farahani2022deep, asadi2024machine} (five classes), and PLCO Breast cancer subtyping classification (three classes) \cite{zhu2013prostate}. For each dataset where patient identifiers were provided, all patient WSIs were aggregated into one set of vectors for input to ABMIL.

\textbf{Survival}
We evaluated the proposed method on five different survival tasks: Breast Invasive Carcinoma (BRCA) from TCGA and PLCO, Non-Small Cell Lung Carcinoma (NSCLC) from PLCO, Uterine Corpus Endometrial Carcinoma (UCEC) from TCGA, Prostate cancer progression-free survival (PRAD) from TCGA.

\textbf{Prostate Cancer Biochemical Recurrence}
We evaluated the proposed solution on one prostate cancer biochemical recurrence dataset of whole-mount radical prostatectomies obtained from our instutition.

\textbf{Gene Biomarker Prediction}
Following the dataset curation of Kather et al. \cite{kather2020pan}, we evaluated our method on six gene-alteration predictions tasks. For colorectal cancer, we combined TCGA-COAD and TCGA-READ (TCGA-CRC) to predict occurence of driver mutations in KRAS, BRAF, and TP53. For breast cancer, we used TCGA-BRCA to predict receptor status (HER2, PR, ER).

\subsection{Baselines and Ablations}
We compared our approach against three baseline vision encoders: UNI, TANGLE (pancancer-trained) (Section \ref{sec:related_work}), CONCH \cite{lu2024visual} and Prov-GigaPath \cite{xu2024whole}. 

Unlike the other methods, TANGLE provides a direct slide-level vector representation, eliminating the need for patch-level aggregation. For TANGLE, we evaluated the model with linear probing on the slide-level features to assess its performance on downstream tasks.

Alongside the baseline comparisons, we conducted an ablation study to evaluate different strategies for routing data experts (Sections \ref{sec:slide_repr}, \ref{sec:ablations}). We further examined the impact of varying the number of clusters in both stages of the two-step clustering procedure (Section \ref{sec:ablations}), as well as a simplified variant of SPADE using one-step K-means clustering to assess the contribution of the additional clustering step (see \ref{app:add_exps}).

To thoroughly evaluate SPADE, we conducted a comparison with a single-expert model (SPADE-0) across subtyping, survival, and BCR prediction tasks (see \ref{app:add_exps_sexp}).

In addition, we explored the effect of the underlying foundation model by replacing the UNI encoder with the Virchow2 foundation model \cite{zimmermann2024virchow2} within SPADE and compared its performance to both standard Virchow2 features and a single-expert version of SPADE trained under the same settings (see \ref{app:add_exps_virch}).

\subsection{Implementation Details}
We used publicly available implementations and pre-trained weights for all baseline foundational models in our study. The preprocessing pipelines were adopted from the official repositories of each model, ensuring consistency with their original training procedures. No additional fine-tuning or retraining was performed on these models beyond their publicly released checkpoints.

We trained the classification ABMIL from scratch using the same set of hyperparameters across all downstream experiments (see \ref{sec:expr_details_sup}).

\subsection{Results}
\label{sec:results}

\subsubsection{Cancer Subtyping}

The SPADE framework consistently outperforms baseline models across a range of cancer subtyping tasks, showing clear advantages in both AUROC and F1-score metrics. Notably, SPADE achieves the highest AUC across seven out of eight tasks and secures the highest F1 score for seven of the eight tasks as well (Table \ref{table:subt_res}). CONCH features led to higher AUC for the breast (TCGA-BRCA) subtyping task. UNI features resulted in a higher F1 score for the ovarian cancer suptyping task (UBC-OCEAN).
The comparison with TANGLE's performance highlights the distinct advantage of spatial transcriptomics over Bulk RNA-seq. Using Bulk RNA-seq may limit model expressibility, likely due to the loss of heterogeneity and the lack of one-to-one alignment between expression and imaging data.  


\begin{table}[H]
\centering
\resizebox{\textwidth}{!}{
\begin{tabular}{l|cccc|cccc|cc|cc|cccc}
            & \multicolumn{4}{c|}{Lung}                                                          & \multicolumn{4}{c|}{Prostate}                                                                                                  & \multicolumn{2}{c|}{\begin{tabular}[c]{@{}c@{}}Lymph\\ Nodes\end{tabular}} & \multicolumn{2}{c|}{Ovarian}                                              & \multicolumn{4}{c}{Breast}                                                                                                    \\ \hline
            & \multicolumn{2}{c|}{PLCO}                          & \multicolumn{2}{c|}{CPTAC}    & \multicolumn{2}{c|}{PANDA}                         & \multicolumn{2}{c|}{\begin{tabular}[c]{@{}c@{}}TCGA-\\ PRAD\end{tabular}} & \multicolumn{2}{c|}{CAM16}                                                 & \multicolumn{2}{c|}{\begin{tabular}[c]{@{}c@{}}UBC-\\ OCEAN\end{tabular}} & \multicolumn{2}{c|}{PLCO}                          & \multicolumn{2}{c}{\begin{tabular}[c]{@{}c@{}}TCGA-\\ BRCA\end{tabular}} \\ \hline
            & AUC           & \multicolumn{1}{c|}{F1}            & AUC           & F1            & AUC           & \multicolumn{1}{c|}{F1}            & AUC                                 & F1                                  & AUC                                  & F1                                  & AUC                                 & F1                                  & AUC           & \multicolumn{1}{c|}{F1}            & AUC                                 & F1                                 \\ \hline
TANGLE      & 95.9          & \multicolumn{1}{r|}{{ 91.1}}    & 96.7          & 89.7          & 88.1          & \multicolumn{1}{c|}{57.4}          & -                                   & -                                   & 74.2                                 & 57.9                                & {\ul 96.4}                          & 81.5                                & 76.4          & \multicolumn{1}{c|}{{\ul 55.3}}    & -                                   & -                                  \\ \hline
GigaPath    & 97.0          & \multicolumn{1}{c|}{90.1}          & {\ul 97.1}    & {\ul 89.9}    & 72.5          & \multicolumn{1}{c|}{33.9}          & 75.3                                & 38.7                                & 93.0                                 & {\ul 88.3}                          & 94.2                                & 77.9                                & 69.6          & \multicolumn{1}{c|}{50.4}          & 69.4                                & 63.9                               \\ \hline
UNI+ABMIL   & 97.1          & \multicolumn{1}{c|}{89.3}          & 96.9          & 88.8          & {\ul 93.2}    & \multicolumn{1}{c|}{{\ul 66.8}}    & {\ul 81.0}                          & {\ul 52.7}                          & {\ul 96.0}                           & 81.7                                & \textbf{97.8}                       & \textbf{89.1}                       & {\ul 79.7}    & \multicolumn{1}{c|}{55.0}          & 97.5                                & {\ul 97.1}                         \\
CONCH+ABMIL & 97.0          & \multicolumn{1}{c|}{\ul{91.3}} & 96.7          & 88.8          & 90.9          & \multicolumn{1}{c|}{62.3}          & 78.9                                & 47.9                                & 81.8                                 & 74.4                                & 96.5                                & 82.9                                & 76.7          & \multicolumn{1}{c|}{51.6}          & \textbf{98.3}                       & 96.7                               \\
SPADE+ABMIL & \textbf{98.5} & \multicolumn{1}{c|}{\textbf{95.5}}          & \textbf{97.9} & \textbf{92.5} & \textbf{94.9} & \multicolumn{1}{c|}{\textbf{74.7}} & \textbf{82.1}                       & \textbf{56.9}                       & \textbf{98.6}                        & \textbf{89.1}                       & \textbf{97.8}                       & {\ul 87.1}                          & \textbf{82.1} & \multicolumn{1}{c|}{\textbf{57.3}} & {\ul 97.7}                          & \textbf{98.9}                      \\ \hline
\end{tabular}
}
\caption{\textbf{Subtyping prediction} results of SPADE and baselines for eight different subtyping tasks. Metrics AUC and F1 are shown as percentages.  
The best performance is in bold, and the second best is underlined. TANGLE could not be applied to TCGA datasets since they were used to train the model.}
\label{table:subt_res}
\end{table}

\subsubsection{Survival Prediction and Prostate BCR Analysis}
We used the concordance index (c-index) for survival and prostate cancer biochemical recurrence (BCR) evaluation (Table \ref{table:surv_res}). To overcome the limitations of overall survival, which includes deaths unrelated to cancer, we utilize disease-specific survival (DSS) for the four survival tasks \cite{liu2018integrated}.

SPADE outperforms all baseline models, achieving the highest c-index for four out of the five survival tasks and the BCR prediction task. In the TCGA-BRCA survival task, UNI surpasses SPADE, but SPADE still achieves the second-highest performance. SPADE outperforms TANGLE on all datasets, demonstrating the critical importance of spatial context in modeling disease progression, which is achieved by incorporating ST data.

\begin{table}[H]
\centering
\resizebox{\textwidth}{!}{
\begin{tabular}{l|cc|c|c|cc}
\multicolumn{1}{c|}{} & \multicolumn{2}{c|}{Prostate} & Lung          & Uterine       & \multicolumn{2}{c}{Breast}    \\ \hline
\multicolumn{1}{c|}{} & UCLA (BCR)         & TCGA-PRAD       & PLCO          & TCGA-UCEC     & TCGA-BRCA     & PLCO          \\ \hline
                      & c-index         & c-index         & c-index       & c-index       & c-index       & c-index       \\ \hline
TANGLE                & 60.3            & -               & 48.6          & -             & -             & {\ul 57.0}    \\ \hline
GigaPath              & {\ul 66.4}      & 53.0            & {\ul 59.5}    & {\ul 69.1}    & 65.5          & 56.0          \\ \hline
UNI+ABMIL             & 58.8            & {\ul 56.5}      & 57.8          & 64.6          & \textbf{74.1} & 56.0          \\
CONCH+ABMIL           & 60.3            & 55.9            & 35.8          & 64.5          & 64.5          & 55.5          \\
SPADE+ABMIL           & \textbf{67.0}   & \textbf{59.4}   & \textbf{64.7} & \textbf{72.4} & {\ul 71.8}    & \textbf{59.5} \\ \hline
\end{tabular}
}
\caption{\textbf{Survival and BCR prediction} results of SPADE and baselines for six different survival tasks. c-index values are shown as percentages.  
The best performance is in bold, and the second best is underlined. TANGLE could not be applied to TCGA datasets since they were used to train the model.}
\label{table:surv_res}
\end{table}
\subsubsection{Gene Biomarker Prediction}
Predicting gene alterations and receptor status from histology slides alone is challenging, and the datasets often suffer from high imbalance. Across the six tasks evaluated, SPADE achieves the highest AUROC on five of the tasks (Table \ref{table:biomarker}). The sole exception is TCGA-CRC BRAF mutation prediction, where CONCH performs slightly better. SPADE also yields the top F1 performance on ER, PR, and TP53 tasks, with CONCH performing better on HER2, GigaPath on KRAS, and TANGLE on BRAF. Averaged over all tasks, SPADE attains the highest mean AUROC (79.4) and mean F1 (64.6).

\begin{table}[H]
\centering
\resizebox{\textwidth}{!}{
\begin{tabular}{l|cccccc|cccccc}
            & \multicolumn{6}{c|}{TCGA-BRCA}                                                                                                           & \multicolumn{6}{c}{TCGA-CRC}                                                                                                            \\ \hline
            & \multicolumn{2}{c|}{HER2}                          & \multicolumn{2}{c|}{ER}                             & \multicolumn{2}{c|}{PR}       & \multicolumn{2}{c|}{KRAS}                          & \multicolumn{2}{c|}{BRAF}                          & \multicolumn{2}{c}{TP53}      \\ \hline
            & AUC           & \multicolumn{1}{c|}{F1}            & AUC           & \multicolumn{1}{c|}{F1}             & AUC           & F1            & AUC           & \multicolumn{1}{c|}{F1}            & AUC           & \multicolumn{1}{c|}{F1}            & AUC           & F1            \\ \hline
TANGLE      & -          & \multicolumn{1}{r|}{-}         & -          & \multicolumn{1}{c|}{-}          & -         & -        & -          & \multicolumn{1}{c|}{{-}}    & -         & \multicolumn{1}{c|}{-} & -         & -         \\ \hline
GigaPath    & {\ul 73.3}    & \multicolumn{1}{c|}{26.82}         & 89.9          & \multicolumn{1}{c|}{93.09}          & 81.3          & 83.27         & {\ul 59.6}    & \multicolumn{1}{c|}{\textbf{51.6}} & 73.1          & \multicolumn{1}{c|}{14.7}          & 81.0          & 84.0          \\ \hline
UNI+ABMIL   & 72.2          & \multicolumn{1}{c|}{30.03}         & {\ul 91.47}   & \multicolumn{1}{c|}{{\ul 96.06}}    & 84.0          & 88.4          & 56.9          & \multicolumn{1}{c|}{44.7}          & 67.0          & \multicolumn{1}{c|}{\ul 22.3}          & {\ul 82.6}    & {\ul 84.4}    \\
CONCH+ABMIL & 68.3          & \multicolumn{1}{c|}{\textbf{39.6}} & 88.67         & \multicolumn{1}{c|}{94.88}          & {\ul 86.5}    & {\ul 91.7}    & 56.1          & \multicolumn{1}{c|}{44.5}          & \textbf{80.0} & \multicolumn{1}{c|}{{\textbf{24.3}}}    & 75.9          & 79.6          \\
SPADE+ABMIL & \textbf{75.3} & \multicolumn{1}{c|}{{\ul 39.0}}    & \textbf{93.1} & \multicolumn{1}{c|}{\textbf{97.76}} & \textbf{87.7} & \textbf{93.0} & \textbf{62.2} & \multicolumn{1}{c|}{\ul 48.7}          & {\ul 74.6}    & \multicolumn{1}{c|}{22.2}          & \textbf{83.5} & \textbf{87.0} \\ \hline
\end{tabular}
}
\caption{\textbf{Molecular biomarker prediction results on TCGA cohorts.} 
Comparison of SPADE with baseline methods for predicting receptor status in TCGA-BRCA (HER2, ER, PR) and mutational status in TCGA-CRC (KRAS, BRAF, TP53). Results are reported in terms of AUC and F1 score (\%). The best performance is in bold, and the second best is underlined. TANGLE could not be applied to TCGA datasets since they were used to train the model.}
\label{table:biomarker}
\end{table}

\subsubsection{Ablation Studies}
\label{sec:ablations}
For the data experts' routing ablation studies, there is varying performance between different routing techniques (Table \ref{table:subt_res_ablation}, \ref{ablation_surv}). For most subtyping datasets, $p_{2}$ (averaging) and $p_{3}$ (distance weighting) routing methods perform better than $p_{1}$, which simply uses the expert of the closest centroid. This suggests that incorporating additional model input from other experts through ensembling provides a performance advantage, potentially enabling more complex representations of the WSI. 

\begin{table}[H]
\centering
\resizebox{\textwidth}{!}{
\begin{tabular}{l|cccc|cc|cc|cc|cc}
              & \multicolumn{4}{c|}{Lung}                                     & \multicolumn{2}{c|}{Prostate} & \multicolumn{2}{c|}{Lymph Nodes} & \multicolumn{2}{c|}{Ovarian}   & \multicolumn{2}{c}{Breast}    \\ \hline
              & \multicolumn{2}{c}{PLCO}      & \multicolumn{2}{c|}{CPTAC}    & \multicolumn{2}{c|}{PANDA}    & \multicolumn{2}{c|}{CAM16}       & \multicolumn{2}{c|}{UBC-OCEAN} & \multicolumn{2}{c}{PLCO}      \\ \hline
              & AUC           & F1            & AUC           & F1            & AUC           & F1            & AUC             & F1             & AUC            & F1            & AUC           & F1            \\ \hline
$p_{1}$ & {\ul 97.9}    & 85.0          & 97.1          & {\ul 91.1}    & 94.4          & {\ul 71.6}    & {\ul 97.1}      & {\ul 87.5}     & 96.8           & 76.9          & {\ul 79.3}    & 51.4          \\
$p_{2}$ & 97.6          & {\ul 95.2} & {\ul 97.7}    & 90.4          & 94.0          & 71.1          & 93.6            & 77.6           & {\ul 97.2}     & {\ul 80.0}    & 76.6          & {\ul 55.6}    \\
$p_{3}$ & \textbf{98.5} & {\textbf{95.5}}    & \textbf{97.9} & \textbf{92.5} & \textbf{94.9} & \textbf{74.7} & \textbf{98.6}   & \textbf{89.1}  & \textbf{97.8}  & \textbf{87.1} & \textbf{82.1} & \textbf{57.3} \\ \hline
\end{tabular}
}
\caption{\textbf{
Ablation study of the data experts' routing method for subtyping.} $p_{1}$  corresponds to the case where the expert associated with the closest centroid is selected for each data point. $p_{2}$ corresponds to the case where the outputs of all experts are averaged to produce a final prediction. $p_{3}$ corresponds to the case where the expert selection is inversely proportional to the distance  (Section \ref{sec:slide_repr}).
Metrics AUC and F1 are shown as percentages.  
The best performance is in bold, and the second best is underlined. 
}
\label{table:subt_res_ablation}
\end{table}

\begin{table}[H]
\centering
\resizebox{\textwidth}{!}{
\begin{tabular}{l|cc|c|c|cc}
\multicolumn{1}{c|}{} & \multicolumn{2}{c|}{Prostate} & Lung          & Uterine       & \multicolumn{2}{c}{Breast}                         \\ \hline
\multicolumn{1}{c|}{} & UCLA (BCR)        & TCGA-PRAD       & PLCO          & TCGA-UCEC     & \multicolumn{1}{c|}{TCGA-BRCA}     & PLCO          \\ \hline
                      & c-index         & c-index         & c-index       & c-index       & \multicolumn{1}{c|}{c-index}       & c-index       \\ \hline
$p_{1}$         & 63.7            & 57.6            & 56.6          & 70.7          & \multicolumn{1}{c|}{68.3}          & 56.6          \\
$p_{2}$        & {\ul 66.8}      & {\ul 59.0}      & {\ul 57.8}    & {\ul 71.5}    & \multicolumn{1}{c|}{{\ul 69.5}}    & {\ul 57.0}    \\
$p_{3}$        & \textbf{67.0}   & \textbf{59.4}   & \textbf{64.7} & \textbf{72.4} & \multicolumn{1}{c|}{\textbf{71.8}} & \textbf{59.5} \\ \hline
\end{tabular}
}
\caption{\textbf{
The ablation study of the data experts' routing method for survival and BCR.} $p_{1}$  corresponds to the case where the expert associated with the closest centroid is selected for each data point. $p_{2}$ corresponds to the case where the outputs of all experts are averaged to produce a final prediction. $p_{3}$ corresponds to the case where the expert selection is inversely proportional to the distance  (Section \ref{sec:slide_repr}).
The best performance is in bold, and the second best is underlined. c-index values are shown as percentages.  
}
\label{ablation_surv}
\end{table}

Our ablation study on the two-step clustering procedure revealed that varying the number of clusters at each stage ($k_1$ and $k_2$) can influence model performance, but the improvement varies across different datasets and cancer types (see Table \ref{ablation_k}, \ref{ablation_k_surv}). For the subtyping task, the configuration with $k_1=16, k_2=16$ achieved the highest AUC in five datasets, demonstrating the effectiveness of balanced hierarchical clustering. Increasing the second-stage clustering granularity ($k_2=32$) improved F1 scores in several cases, such as ovarian, suggesting that a finer-grained latent space can enhance classification precision, but it also requires double the computational resources or double the runtime. Reducing the second-stage clustering granularity ($k_2=8$) led to mixed results: while it improved F1 in some datasets, it generally resulted in lower AUC scores.

\begin{table}[H]
\centering
\resizebox{\textwidth}{!}{
\begin{tabular}{l|cccc|cccc|cc|cc|cccc}
                             & \multicolumn{4}{c|}{Lung}                                      & \multicolumn{4}{c|}{Prostate}                                                                             & \multicolumn{2}{c|}{Lymph Nodes} & \multicolumn{2}{c|}{Ovarian}   & \multicolumn{4}{c}{Breast}                                                                               \\ \hline
                             & \multicolumn{2}{c}{PLCO}       & \multicolumn{2}{c|}{CPTAC}    & \multicolumn{2}{c}{PANDA}     & \multicolumn{2}{c|}{\begin{tabular}[c]{@{}c@{}}TCGA-\\ PRAD\end{tabular}} & \multicolumn{2}{c|}{CAM16}       & \multicolumn{2}{c|}{UBC-OCEAN} & \multicolumn{2}{c}{PLCO}      & \multicolumn{2}{c}{\begin{tabular}[c]{@{}c@{}}TCGA-\\ BRCA\end{tabular}} \\ \hline
                             & AUC           & F1             & AUC           & F1            & AUC           & F1            & AUC                                 & F1                                  & AUC             & F1             & AUC            & F1            & AUC           & F1            & AUC                                 & F1                                 \\ \hline
$16(k_{1}), 16(k_{2})$ & \textbf{98.5} & \textbf{95.5}           & {\ul 97.9}    & \textbf{92.5} & \textbf{94.8} & {\ul 74.7}    & {\ul 82.1}                          & \textbf{56.9}                       & \textbf{98.7}   & 89.1           & \textbf{97.8}  & {\ul 87.1}    & \textbf{82.1} & 57.3          & {\ul 97.7}                          & 99.0                               \\
$16(k_{1}), 8(k_{2})$  & 96.8          & {\ul 91.3}     & 97.3          & 89.6          & 94.4          & \textbf{76.0} & 81.4                                & 49.3                                & 93.6            & 87.3           & {\ul 97.7}     & 85.3          & 80.3          & \textbf{58.6} & 97.4                                & 98.0                               \\
$16(k_{1}), 32(k_{2})$ & {\ul 97.8}    & \textbf{95.5} & \textbf{98.1} & {\ul 90.2}    & {\ul 94.6}    & 71.2          & 81.4                                & {\ul 51.1}                          & \textbf{98.7}   & 90.5           & 97.4           & \textbf{88.0} & 78.6          & 57.6          & 97.5                                & 99.1                               \\
$32(k_{1}), 32(k_{2})$ & 97.2          & 87.8           & 96.9          & 84.6          & 94.1          & 69.7          & \textbf{83.5}                       & 50.8                                & {\ul 98.6}      & \textbf{91.1}  & 96.9           & 78.3          & 77.5          & {\ul 58.4}    & \textbf{97.8}                       & \textbf{99.3}                      \\ \hline
\end{tabular}
}
\caption{\textbf{
Ablation study of the number of experts for each clustering step for subtyping.} Metrics AUC and F1 are shown as percentages.
The best performance is in bold, and the second best is underlined. 
}
\label{ablation_k}
\end{table}

For the survival and BCR, $k_1=16, k_2=16$ achieved the highest c-index for four out of six tasks, demonstrating the effectiveness of balanced clustering. Increasing $k_1$ and  $k_2$  to 32 improved the c-index for TCGA-BRCA while reducing $k_2$ to 8 led to performance drops across all survival tasks.

\begin{table}[H]
\centering
\resizebox{\textwidth}{!}{
\begin{tabular}{l|cc|c|c|cc}
                             & \multicolumn{2}{c|}{Prostate} & Lung          & Uterine       & \multicolumn{2}{c}{Breast}    \\ \hline
                             & UCLA (BCR)         & TCGA-PRAD       & PLCO          & TCGA-UCEC     & TCGA-BRCA     & PLCO          \\ \hline
                             & c-index         & c-index         & c-index       & c-index       & c-index       & c-index       \\ \hline
$16(k_{1}), 16(k_{2})$ & 67.0            & \textbf{59.4}   & \textbf{64.7} & \textbf{72.4} & {\ul 71.8}    & \textbf{59.5} \\
$16(k_{1}), 8(k_{2})$  & 66.2            & 55.7            & 53.2          & 70.4          & 70.8          & 56.1          \\
$16(k_{1}), 32(k_{2})$ & \textbf{70.2}   & {\ul 57.4}      & {\ul 60.7}    & 71.7          & 71.7          & 57.0          \\
$32(k_{1}), 32(k_{2})$ & {\ul 68.9}      & 56.5            & 56.5          & {\ul 72.1}    & \textbf{72.0} & {\ul 57.2}    \\ \hline
\end{tabular}
}
\caption{\textbf{
Ablation study of the number of experts for each clustering step for survival and BCR.} c-index values are shown as percentages.  
The best performance is in bold, and the second best is underlined. 
}
\label{ablation_k_surv}
\end{table}

Further comparisons between SPADE and its single-expert version highlight the significant impact of the mixture of data experts module on the performance, as evidenced by SPADE’s consistently higher AUC and F1 score (see \ref{app:add_exps}). This trend may be attributed to the diverse tissue types represented in the HEST-1k dataset (e.g., brain, heart, lung). A model trained on all organ types might lack the specificity needed to improve performance for each task. Given the high variability of genes present between tissue types, training on data from all organs within a single model may limit its ability to capture the heterogeneity essential for effective subtyping and survival prediction. Additionally, as discussed in Section \ref{sec:mode}, with such diverse tissue types, contrastive learning could rely on easy negative pairings to drive down the loss, resulting in a less expressive latent space.

\subsubsection{Interpretability Analysis}

\begin{figure}[H]
    \centering
    \includegraphics[width=\textwidth]{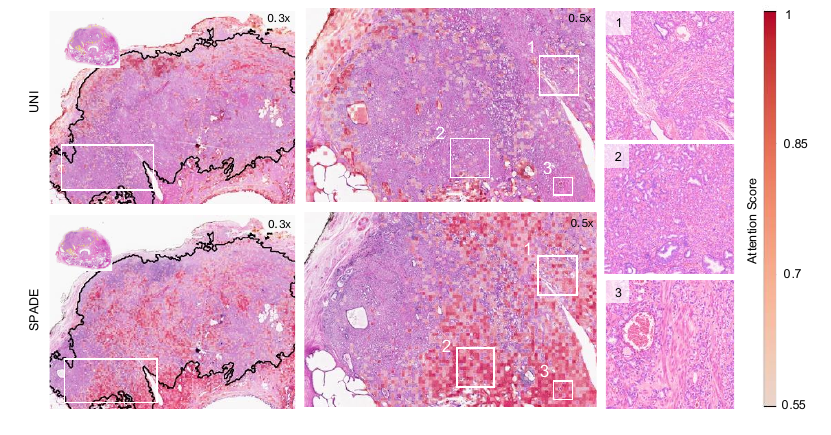} %
    \caption{\textbf{Interpretability of SPADE}. Attention scores for a prostate WSI positive for biochemical recurrence (BCR) are visualized as a heatmap for the UNI and SPADE models. The deeper the red color, the higher attention the model puts on that region of the tissue. The black outline shows the predicted tumor boundary from our internal cancer prediction model for whole-mount prostectomy WSIs. SPADE shows a higher concentration of attention within the tumor boundary compared to UNI, with some areas of focus along the tumor border. Sub-crops are selected to show the patterns with high attention at higher magnifications.}
    \label{fig:attention} 
\end{figure}

We conducted our first interpretability analysis by plotting the attention weights from the ABMIL variants on our in-house UCLA Prostate BCR dataset to better understand the differences between the SPADE and UNI features. In Figure \ref{fig:attention}, we present attention heatmaps, where deeper red indicates higher attention weights, and the predicted cancer mask is shown in yellow. We validated this cancer prediction mask against pathology reports. In panel (a), UNI attention highlights some small regions within the tumor boundary, but much of the attention is diffuse. In contrast, the SPADE framework's attention weights display higher attention regions within the cancer region and a hot zone along the bottom right border of the tumor mask.

The heatmap depicts which regions of the WSI had the highest attention for a case from the UCLA BCR task. The attention heatmap produced by SPADE highlights areas predominantly showing poorly formed and compact $4+5$ and $4+3$ tumor patterns with tertiary $5$ according to the Gleason grading system \cite{epstein20162014}, subtle desmoplastic stromal response with slightly cellular fibroblasts exhibiting pale cytoplasm and nuclear enlargement.

\subsubsection{Comparison of TCGA-CRC Attention Heatmaps from SPADE vs. UNI}
\label{sec:crc_att}
To compare SPADE and UNI for gene-related tasks, attention weights from the TCGA-CRC BRAF mutation task were extracted to identify differences in the attentions of each model.

\begin{figure}[H]
    \centering
    \includegraphics[width=\textwidth]{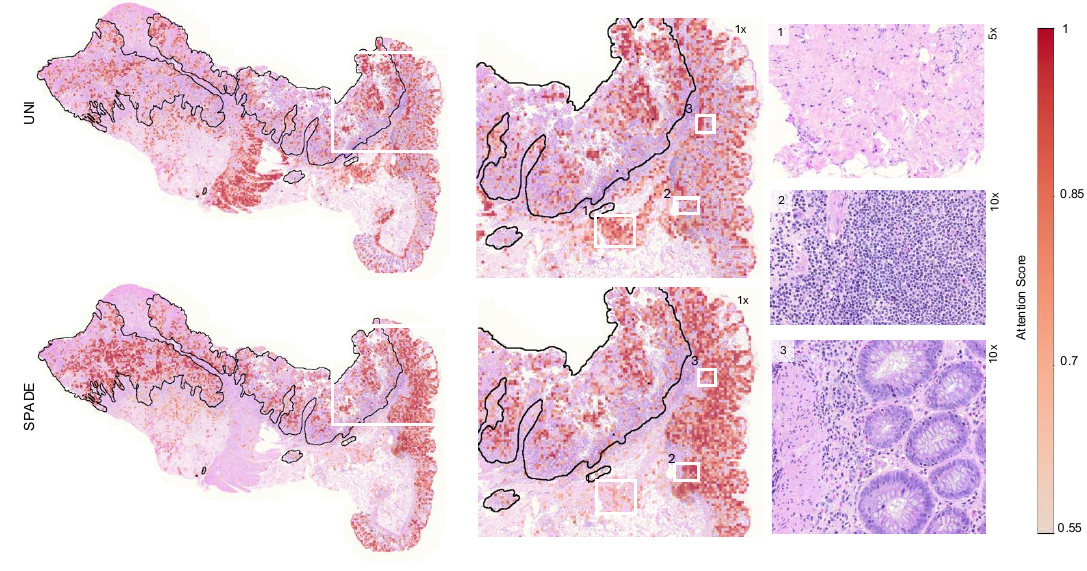} %
    \caption{\textbf{Interpretability of SPADE on TCGA-CRC BRAF Task}. Attention scores for a colorectal cancer WSI are visualized as a heatmap for the UNI and SPADE models. The deeper the red color, the higher attention the model puts on that region of the tissue. Sub-crops are selected to show the patterns with high attention at higher magnifications. A pathologist noted differences between UNI and SPADE heatmaps (see Section \ref{sec:crc_att}). Cancer outline delineated by a pathologist in black.} 
    \label{fig:crc_att3} 
\end{figure}

We created a series of heatmaps similar to our setup for Prostate BCR attention visualization (see Figure \ref{fig:crc_att3}, \ref{fig:crc_att1}, \ref{fig:crc_att2}). Additionally, we extracted these heatmaps at high resolution and created a setup for a board-certified pathologist to overlay the heatmap on tissue at full resolution. This setup allowed the pathologist to zoom in on "hot" regions and examine the underlying tissue patterns. From the heatmaps, the pathologist observed that UNI tended to focus on non-neoplastic areas containing smooth muscle and chronic inflammation, rather than tumor glands. In contrast, the pathologist noted that SPADE focused on areas with high grade dysplasia and invasive carcinoma. In one case SPADE also focused on regions with elongated nuclei ("spindle"-like). While spindle-cell morphology is not a canonical pattern of BRAF in colorectal cancer samples, spindle-cell patterns have been described in non-CRC BRAF-altered neoplasms such as melanoma \cite{suurmeijer2018novel}. In another case, the pathologist noted that UNI's attention pattern appeared more random, focusing on necrotic regions rather than following the glandular architecture. Overall, in the six cases reviewed by our pathologist, UNI focused primarily on smooth muscle tissue and regions without tumor, whereas SPADE concentrated more on tumor regions.

\begin{figure}[H]
    \centering
    \includegraphics[width=\textwidth]{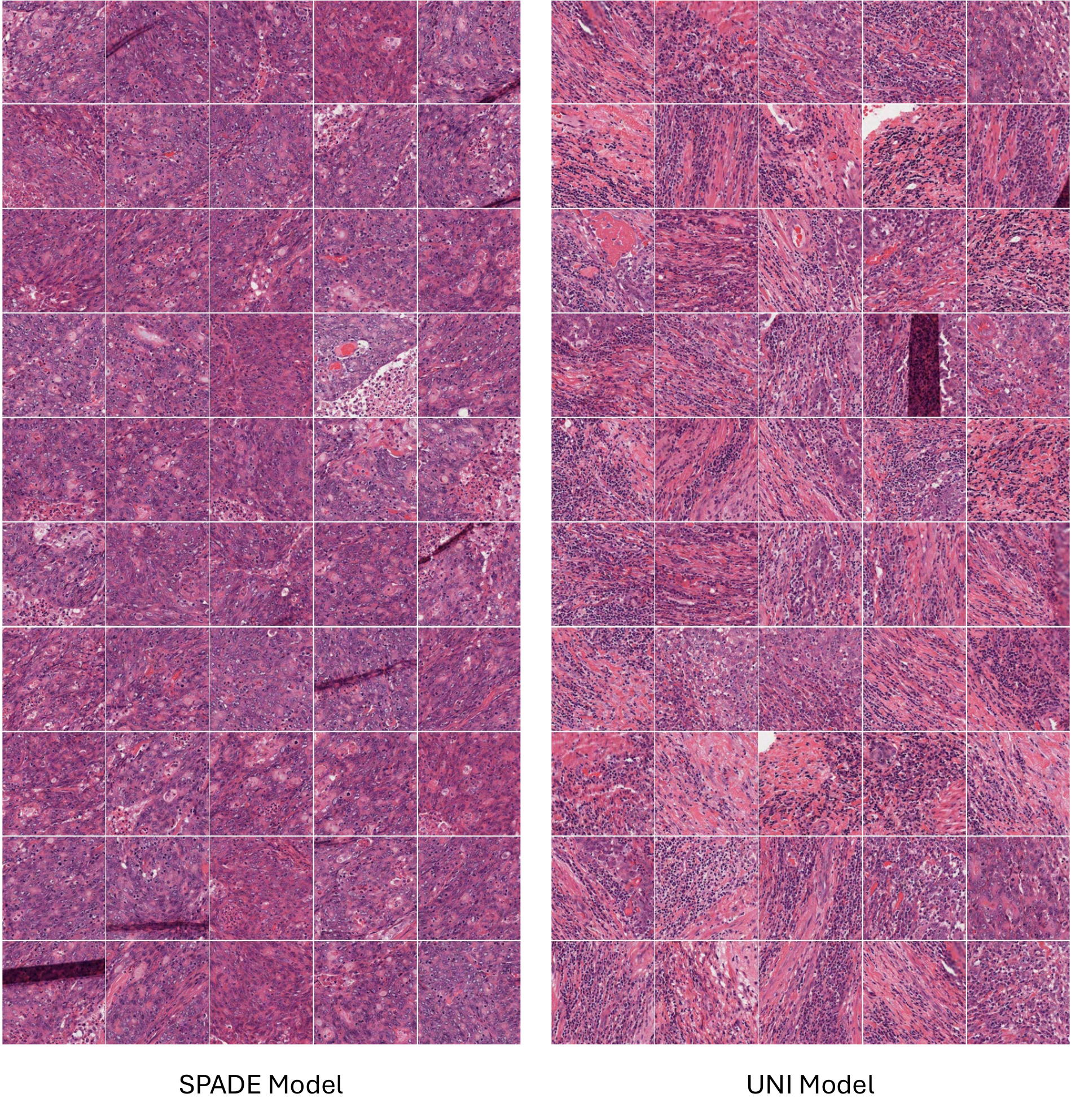} %
    \caption{\textbf{Highest Attention Patches SPADE vs UNI}. We used the attention scores to extract the 50 highest-scoring attention patches from our model, SPADE, and compared them to the 50 highest-scoring attention patches from UNI. We then provided the resulting plot to a pathologist for review, with model names withheld. The pathologist noted differences between UNI and SPADE highest attention patches (see Section \ref{sec:crc_att}).}
    \label{fig:crc_path_review} 
\end{figure}

Next, we extracted the 50-highest attention patches from five different cases at 20x magnification for both SPADE and UNI (see Figure \ref{fig:crc_path_review}). We then provided patches to the pathologist, grouped by model and case, with model names obscured. Across the five cases, SPADE generally highlighted malignant glands, often with high-grade or cribriform architecture. UNI, in contrast, frequently highlighted lower-grade or fewer glands in its top 50 highest attention patches, and often emphasized stroma, fibrotic tissue, or inflammation rather than the tumor. In several cases, UNI identifies tiles with little or no carcinoma. The clinical impression suggests that SPADE is more tumor-focused, while UNI shifts attention to stromal or non-cancerous regions.

\section{Discussion}

Our study demonstrates that training on paired spatial transcriptomics and histology data substantially enhances the expressiveness of deep histology features across diverse cancer tasks. By leveraging gene expression data during contrastive pretraining, our framework improved performance on 17 out of 20 downstream tasks, including subtyping, survival, and biomarker prediction. These results highlight the value of multimodal supervision in building more robust histopathology foundation models.

Our method introduces a two-step clustering procedure coupled with a mixture-of-data-experts contrastive learning approach, enabling effective hard-negative mining for multi-organ histopathology representation learning. This training strategy enables learning a representation space that does not rely on trivial differences between histology tissues. Importantly, this approach can be extended to any histological dataset with multiple diseases and multiple organs to improve contrastive pretraining.

While we trained on the largest and most diverse public Visium dataset available, the sample size of 515 remains small compared to large digital pathology datasets. However, given the cost of acquiring spatial transcriptomics data and the relative novelty of the technique, it may be some time before datasets of comparable size to those of digital pathology can be assembled. Moreover, spatial transcriptomics gene expression data often suffer from high dropout rates, with many genes exhibiting zero expression levels and other noise-related issues. Although we attempted to mitigate this issue by clustering in the image space and using it to supervise our data experts, incorporating additional supervision using the gene expression space could enhance model robustness and leverage this data type even more. 

Recent advances in spatial transcriptomics foundation models provide an opportunity to utilize pretrained representations on the gene expression side rather than relying solely on raw expression profiles. These models learn biologically meaningful latent spaces from large-scale spatial transcriptomics datasets, capturing both spatial context and transcriptional co-expression patterns that could complement histology-derived features. Integrating such pretrained embeddings (e.g., stFormer \cite{cao2024stformer}, or SToFM \cite{zhao2025stofm}) into multimodal pretraining could further improve alignment between morphological and molecular modalities.

HEST-1k contains various types of spatial transcriptomics data, spanning in-situ imaging techniques (e.g., Xenium) and Next-Generation sequencing techniques (e.g., Visium). In our study, we chose to use only the Visium data, as it provides the whole-genome transcriptome, enabling us to identify sets of highly variable genes shared across tissue types without being restricted by organ-specific panels. In future work, we plan to incorporate other spatial transcriptomics datasets and further expand our dataset. Additionally, the 2-step K-means clustering procedure is sensitive to the choice of K, which may vary in optimality across downstream tasks. We optimized our choice of K using the within-cluster sum of squares, but additional clustering optimization with adaptive or hierarchical clustering methods could further enhance performance.

The SPADE model weights and code are publicly released to facilitate open-source use (\href{https://github.com/uclabair/SPADE}{https://github.com/uclabair/SPADE}). Researchers can apply our pretrained model to their own histology datasets to obtain transcriptomics-enhanced multimodal representations.

\section{Conclusions}
\label{sec:discussion}
Based on the performance across our 20 downstream tasks, the SPADE framework has been shown to be an expressive feature encoder using spatial transcriptomics data. Utilization of HEST-1k allowed our model to learn robust, biologically relevant embeddings by aligning histological image features with spatially resolved gene expression patterns, which set a strong foundation for downstream pathology analysis tasks. Our improved performance over the single expert version of our framework highlights the benefit of SPADE's mixture of experts approach. Additionally, our results show significant improvement over TANGLE, which relies on bulk sequencing data, underscoring the value of spatial gene profiling. SPADE's ability to utilize spatial context allows our model to capture localized differences in gene expression that are crucial for accurate disease subtyping and survival prediction.

\section{Funding}
The authors would like to acknowledge support from the National Cancer Institute of the National Institutes of Health under Award Number \allowbreak R01CA279666 and National Library Of Medicine of the National Institutes of Health under Award Number T15LM013976.

\section{ Declaration of generative AI and AI-assisted technologies in the manuscript preparation process}
During the preparation of this work the authors used ChatGPT and Gemini in order to polish writing. After using this tool/service, the authors reviewed and edited the content as needed and take full responsibility for the content of the published article.

\appendix

\section{Datasets}
We provide descriptions of the datasets
used for the evaluation of SPADE.

\noindent\textbf{NSCLC} For the non-small cell lung carcinoma (NSCLC) subtyping task, we utilize H\&E WSIs from the PLCO and CPTAC datasets to classify cases of lung adenocarcinoma (LUAD) and lung squamous cell carcinoma (LUSC). The PLCO cohort contains 589 slides from 274 patients (LUAD: 171, LUSC: 103).
The CPTAC cohort contains 1331 slides from 427 patients (LUAD: 219, LUSC: 208). We perform a patient-level, label-stratified split of both cohorts into training, validation, and test sets with a 70:10:20 ratio. Performance was evaluated using AUCROC and F1-Score.

We also ran a lung cancer survival task using the same PLCO lung cohort. Mortality status from the PLCO study data is used to determine survival, censorship, and duration of survival for this dataset. Out of 274 participants, 213 survived. We performed a patient-level, label-stratified split into training, validation, and test sets with a 70:10:20 ratio. Performance was evaluated using the c-index.

\noindent\textbf{PANDA} For the prostate cancer ISUP grading task, we utilized 10,616 core needle biopsy cases collected from Karolinska Institute and Radboud University Medical Center. All biopsy WSIs were obtained from the Prostate Cancer Grade Assessment (PANDA) challenge. Each biopsy is assigned an ISUP grade on a scale from 1 to 5 if cancerous, while benign biopsies receive a label of 0, resulting in a six-class classification problem. The class distribution is as follows: class 0  (2892), class 1  (2666), class 2  (1343), class 3  (1242), class 4  (1249), and class 5 (1224). Patient identification information was not provided in the dataset, so patient-level splits could not be created.  Instead, we perform a label-stratified split into training, validation, and test sets with a 70:10:20 ratio. Performance was evaluated using AUCROC and macro F1-Score.

\noindent\textbf{Camelyon16} We used the CAMELYON16 dataset, which includes 397 WSIs from sentinel lymph node biopsies, to detect metastases in breast cancer patients. Metastases are categorized into three classes—'negative' (237), 'micro' (80), and 'macro' (80)—making this a three-class classification problem. Patient identification information was not provided in the dataset, so patient-level splits could not be created. Instead, we perform a label-stratified split into training, validation, and test sets with a 70:10:20 ratio. Performance was evaluated using AUCROC and macro F1-Score.

\noindent\textbf{UBC-OCEAN} For ovarian cancer subtype classification, we utilized 513 biopsy samples from the UBC ovarian cancer subtype classification and outlier detection (UBC-OCEAN) competition. Each sample is assigned one out of five histological subtypes of ovarian cancer: class 0 (217), class 1 (42), class 2 (119),  class 3 (94), and class 4 (41). Patient identification information was not provided in the dataset, so patient-level splits could not be created. Instead, we perform a label-stratified split into training, validation, and test sets with a 70:10:20 ratio. Performance was evaluated using AUCROC and macro F1-Score.

\noindent\textbf{Breast} For the breast cancer subtyping task, we used the PLCO Breast dataset. The PLCO cohort consists of 1768 slides from 867 patients. Each sample is categorized into one of three classes - 'lobular' (104), 'ductal' (657), and 'other' (106). We perform patient-level, label-stratified split into training, validation, and test sets with a 70:10:20 ratio. Performance was evaluated using AUCROC and macro F1-Score.

We use the PLCO Breast and TCGA-BRCA datasets for breast cancer survival tasks. Mortality status from the PLCO study data is used to determine survival, censorship, and duration of survival. Out of 867 participants, 552 survived. The TCGA-BRCA dataset consists of 1267 WSIs spanning 1042 patients, with 963 surviving based on disease-specific survival (DSS) status. We perform a patient-level, label-stratified split for both datasets into training, validation, and test sets with a 70:10:20 ratio. Performance was evaluated using the c-index.

We also use the TCGA-BRCA dataset to predict ER, PR, and HER2 receptor status using labels extracted by Kather et al. \cite{kather2020pan}. Patients were split into 70:10:20 train/validation/test splits at the patient level. Performance was evaluated using AUROC and F1. Cohort sizes and prevalences were: ER: 691 patients (532 positive, 159 negative), PR: 688 (463 positive, 225 negative), and HER2: 680 (105 positive, 576 negative).

\noindent\textbf{Prostate}
For our prostate cancer biochemical recurrence (BCR) task, we use an in-house dataset of radical prostatectomy whole-mount WSIs from UCLA. We also use the TCGA-PRAD dataset, however there are no biochemical recurrence labels provided for this dataset, instead we use progression-free survival parsed by \cite{liu2018integrated}. The in-house dataset spans 161 patients, with one whole-mount WSI selected per patient based on pathology reports. The ratio of patients without BCR versus those with BCR is 135:26. The Gleason grading breakdown of this dataset is: 74 - 3+4 cases, 46 - 4+3 cases, 20 - 4+5 cases, 13 - 3+3 cases, 6 - 4+4 cases, 1 - 3+5 case, and 1 - 5+4 case. The TCGA-PRAD dataset consists of 336 WSIs spanning 322 patients with a progression-free survival rate ratio of 285:37. We perform a patient-level, label-stratified split into training, validation, and test sets with a 70:10:20 ratio for both in-house and TCGA-PRAD datasets. Performance was evaluated using the c-index.

\noindent\textbf{Uterine}
For the uterine cancer survival task, we use the TCGA-UCEC dataset. The dataset consists of 786 images spanning 504 patients, with 445 out of the 504 patients total surviving. We perform a patient-level, label-stratified split into training, validation, and test sets with a 70:10:20 ratio. Performance was evaluated using the c-index.

\noindent\textbf{Colorectal}
For colorectal cancer, we combined the TCGA-READ and TCGA-COAD datasets (TCGA-CRC, $N = 555$) and formed 70:10:20 patient-level splits separately for each target. Labels were extracted from \cite{kather2020pan}. We evaluated performance using AUROC and F1 score. Class counts and prevalences were: KRAS: 202 positive, 353 negative; BRAF: 51 positive, 504 negative; TP53: 292 positive, 263 negative.

\section{Model and training}
\label{sec:expr_details_sup}
For pretraining training, we employ a weight decay of $1 \times 10^{-5}$ and use the AdamW optimizer with a learning rate of $1 \times 10^{-4}$, along with a cosine decay scheduler. For the slide classification experiments, we utilized a cross-entropy loss. We employed early stopping if the validation loss failed to improve over ten consecutive epochs with a total training epochs of 20. For survival prediction experiments, we used negative log-likelihood loss (NLL).

\textbf{ABMIL downstream architecture:} ABMIL architecture used in the downstream experiments consists of three components. First, a 2-layer MLP with 256 or 512 hidden units, layer normalization, ReLu activation, and 0.25 dropout.  This is followed by a gated-attention network consisting of 2-layer
MLP, with Sigmoid and Tanh activation, respectively, and 0.25 dropout. Finally, a post-attention linear classification layer with 256 or 512 hidden units is applied.


\textbf{TANGLE downstream architecture:} For the TANGLE baseline, we employed logistic regression following the open-source instructions provided by the authors.

\section{Clustering}
\label{sec:app_clustering}

For the proposed 2-step clustering, the size of the uniform sample $\mathcal{D}^{`}_{i}$ is 1,500,000 patches. The fine-grained K-means is performed over each $\mathcal{D}^{`}_{i}$ with $k=16$. A second round of clustering is performed on top of previously found clusters $S$ with $k=16$.

\begin{figure}[H]
    \centering
    \includegraphics[scale=0.9]{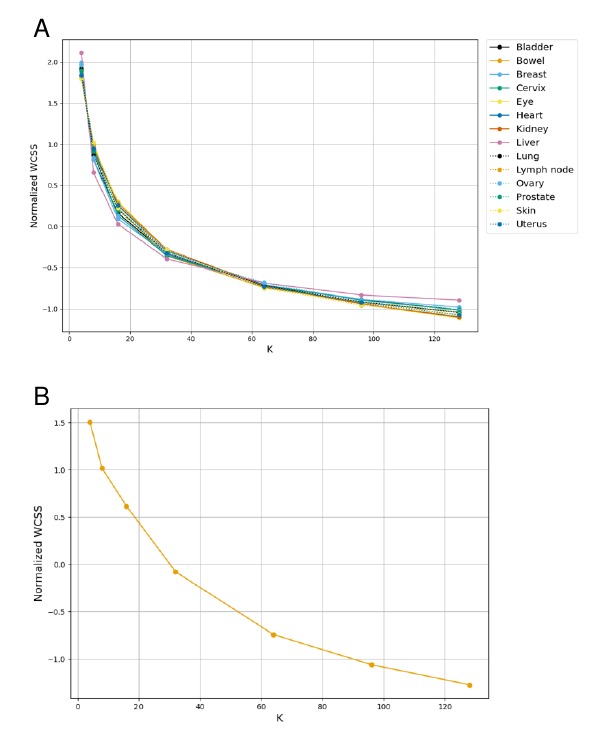} %
    \caption{\textbf{Within-Cluster Sum of Squares (WCSS) Elbow Plots Across Values of K} \textbf{A.} First-level clustering performance evaluated using normalized WCSS across 14 organs. \textbf{B.} Second-level clustering performance evaluated using normalized WCSS.}
    \label{fig:clustering} 
\end{figure}

In the first clustering step, we performed clustering within each tissue type, varying $k$ across a range of values. To determine the optimal number of clusters, we calculated the within-cluster sum of squares (WCSS) and applied the elbow method. The corresponding elbow plot (see Figure \ref{fig:clustering}), suggests that the optimal choice is $k=16$, as improvements in WCSS diminish beyond this point. A secondary drop is observed at $k=32$, which we included in our ablation studies to assess its impact on downstream task performance. 

In the second clustering step, we clustered the centroids obtained prior to form broader, overall clusters. Again, we evaluated values of $k$ using WCSS and generated an elbow plot (see Figure \ref{fig:clustering}). In this case, the distinction between $k=16$ and $k=32$ is less pronounced. However, given the higher computational cost of running 32 experts, we opted for $k=16$ to reduce overall model complexity. This approach allowed us to structure the data efficiently while balancing computational cost and cluster granularity. 

\section{Additional interpretability figures}
\label{sec:add_interp}
\begin{figure}[H]
    \centering
    \includegraphics[width=\textwidth]{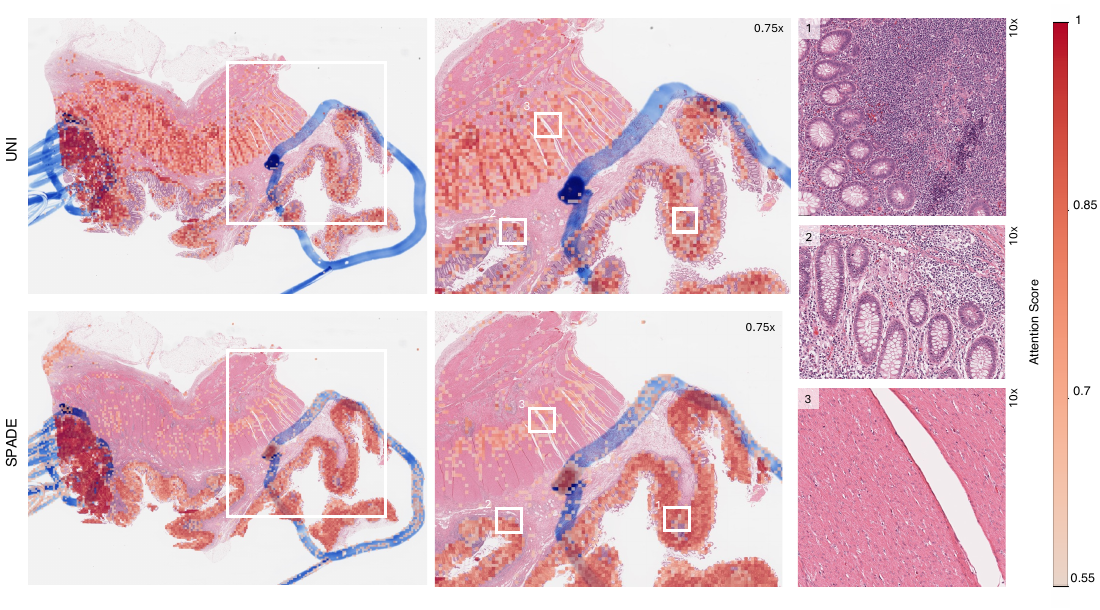} %
    \caption{\textbf{Interpretability of SPADE}. Attention scores for a colorectal cancer WSI are visualized as a heatmap for the UNI and SPADE models. The deeper the red color, the higher attention the model puts on that region of the tissue. Sub-crops are selected to show the patterns with high attention at higher magnifications. A pathologist noted differences between UNI and SPADE heatmaps (see Section \ref{sec:crc_att}).}
    \label{fig:crc_att1} 
\end{figure}

\begin{figure}[H]
    \centering
    \includegraphics[width=\textwidth]{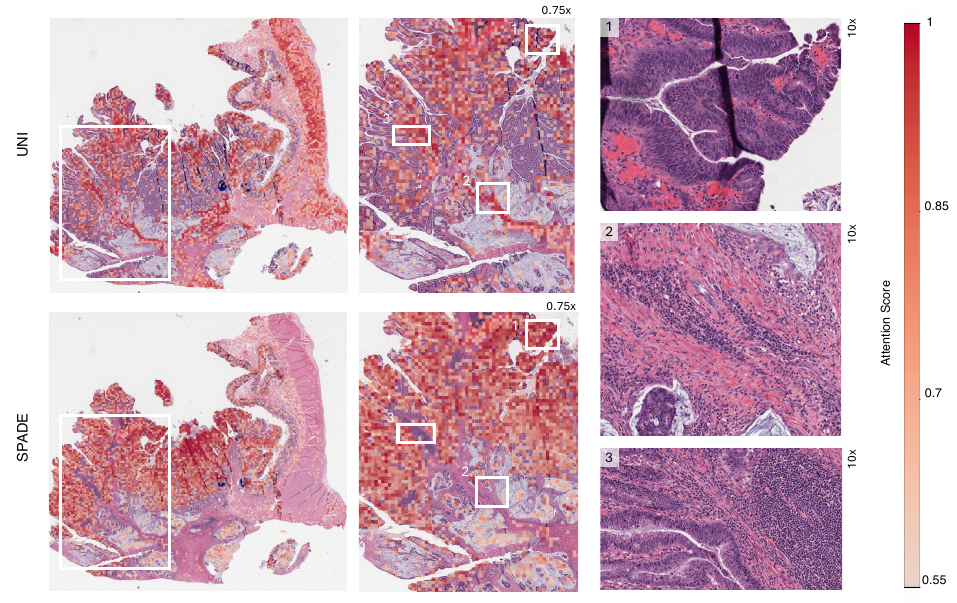} %
    \caption{\textbf{Interpretability of SPADE}. Attention scores for a colorectal cancer WSI are visualized as a heatmap for the UNI and SPADE models. The deeper the red color, the higher attention the model puts on that region of the tissue. Sub-crops are selected to show the patterns with high attention at higher magnifications. A pathologist noted differences between UNI and SPADE heatmaps (see Section \ref{sec:crc_att}).}
    \label{fig:crc_att2} 
\end{figure}

\section{Additional experiments, one-step clustering}
\label{app:add_exps}
We implemented a version of SPADE using clusters generated from the first stage of our two-step clustering algorithm, where each organ-specific cohort was clustered with $k=16$, totaling 240 cluster centroids and, in turn, 240 experts. We report the classification results for two subtyping tasks: Camelyon16 and CPTAC lung.

We observe that data experts trained on cluster centroids obtained with a two-step clustering approach lead to better results compared to a one-step clustering approach (Table \ref{tab:suppl1}). 
The one-step approach results in a reduction in AUC for the CPTAC dataset, with a decrease of 1.9\%, and for the Camelyon dataset, with a decrease of 5.68\%. 
F1 values decreased by 3.03\%  on the CPTAC dataset. For the Camelyon dataset, F1 values dropped by 12.91\%.
This can be explained by the fact that one-step clustering leads to cluster centers that are not sufficiently representative of a specific tissue prototype and, therefore, cannot learn a good representation. 

\begin{table}[H]
\centering
\begin{tabular}{l|cc|cc}
\multicolumn{1}{c|}{} & \multicolumn{2}{c|}{Lung}  & \multicolumn{2}{c}{Lymph Nodes} \\ \hline
\multicolumn{1}{c|}{} & \multicolumn{2}{c|}{CPTAC} & \multicolumn{2}{c}{Camelyon}    \\ \hline
                      & AUC          & F1          & AUC            & F1             \\ \hline
SPADE           & 96.6        & 89.7       & 93.6          & 77.6          \\ \hline

\end{tabular}
\caption{\textbf{One-step clustering for the mixture of experts.} We compare the performance of SPADE with 240 data experts using two subtyping tasks. Metrics AUC and F1 are shown as percentages.}
\label{tab:suppl1}
\end{table}

\section{Additional experiments, single expert}
\label{app:add_exps_sexp}
To further evaluate the effectiveness of SPADE, we conducted an experiment comparing its performance against a single-expert model (SPADE-0) across subtyping, survival and BCR prediction tasks. As shown in Tables \ref{tab:suppl2}, \ref{tab:suppl3}, SPADE consistently outperformed SPADE-0 in most datasets, demonstrating the advantage of incorporating multiple experts tailored to different tissue types.

\begin{table}[H]
\centering
\resizebox{\textwidth}{!}{
\begin{tabular}{l|cccc|cccc|cc|cc|cccc}
                 & \multicolumn{4}{c|}{Lung}                                                          & \multicolumn{4}{c|}{Prostate}                                                                                                  & \multicolumn{2}{c|}{\begin{tabular}[c]{@{}c@{}}Lymph\\ Nodes\end{tabular}} & \multicolumn{2}{c|}{Ovarian}                                              & \multicolumn{4}{c}{Breast}                                                                                                    \\ \hline
                 & \multicolumn{2}{c|}{PLCO}                          & \multicolumn{2}{c|}{CPTAC}    & \multicolumn{2}{c|}{PANDA}                         & \multicolumn{2}{c|}{\begin{tabular}[c]{@{}c@{}}TCGA-\\ PRAD\end{tabular}} & \multicolumn{2}{c|}{CAM16}                                                 & \multicolumn{2}{c|}{\begin{tabular}[c]{@{}c@{}}UBC-\\ OCEAN\end{tabular}} & \multicolumn{2}{c|}{PLCO}                          & \multicolumn{2}{c}{\begin{tabular}[c]{@{}c@{}}TCGA-\\ BRCA\end{tabular}} \\ \hline
                 & AUC           & \multicolumn{1}{c|}{F1}            & AUC           & F1            & AUC           & \multicolumn{1}{c|}{F1}            & AUC                                 & F1                                  & AUC                                  & F1                                  & AUC                                 & F1                                  & AUC           & \multicolumn{1}{c|}{F1}            & AUC                                 & F1                                 \\ \hline
SPADE-0    & 96.1          & \multicolumn{1}{c|}{\textbf{93.3}} & {96.3}    & 87.4          & {93.6}    & \multicolumn{1}{c|}{{ 66.8}}    & {81.5}                          & 39.6                                & {97.1}                           & 85.1                                & 97.1                                & {85.7}                          & {78.7}    & \multicolumn{1}{c|}{\textbf{59.2}} & 97.6                                & 97.0                               \\
SPADE      & \textbf{98.5} & \multicolumn{1}{c|}{{ 87.1}}    & \textbf{97.9} & \textbf{92.5} & \textbf{94.9} & \multicolumn{1}{c|}{\textbf{74.7}} & \textbf{82.1}                       & \textbf{56.9}                       & \textbf{98.6}                        & \textbf{89.1}                       & \textbf{97.8}                       & \textbf{87.1}                       & \textbf{82.1} & \multicolumn{1}{c|}{{57.3}}    & {\textbf {97.7}}                          & \textbf{98.9}                      \\ \hline
\end{tabular}
}
\caption{\textbf{Comparison of mixture-of-experts (SPADE) to single expert (SPADE-0) for subtyping task.} Metrics AUC and F1 are shown as percentages.}
\label{tab:suppl2}
\end{table}

\begin{table}[H]
\centering
\resizebox{\textwidth}{!}{
\begin{tabular}{l|cc|c|c|cc}
\multicolumn{1}{c|}{} & \multicolumn{2}{c|}{Prostate} & Lung           & Uterine        & \multicolumn{2}{c}{Breast}      \\ \hline
\multicolumn{1}{c|}{} & UCLA (BCR)         & TCGA-PRAD       & PLCO           & TCGA-UCEC      & TCGA-BRCA      & PLCO           \\ \hline
                      & c-index         & c-index         & c-index        & c-index        & c-index        & c-index        \\ \hline
SPADE-0           & 0.650           & 0.589           & 0.572          & 0.701          & 0.685          & 0.561          \\
SPADE           & {\textbf{ 0.670}}     & {\textbf{0.594}}     & {\textbf{0.647} }   & \textbf{0.724} & {\textbf 0.718}    & \textbf{0.595} \\ \hline
\end{tabular}
}
\caption{\textbf{Comparison of mixture-of-experts (SPADE) to single expert (SPADE-0) for survival and BCR}. c-index values are shown as percentages.  }
\label{tab:suppl3}
\end{table}

\section{Additional experiments, Virchow2 as a backbone}
\label{app:add_exps_virch}
To assess the generality of SPADE across different histology backbones, we repeated the subtyping experiments using Virchow2 features instead of UNI as the image encoder. Table \ref{tab:virch_subt} compares SPADE(Virchow2) to its single-expert variant SPADE(Virchow2)-0 and to the Virchow2 baseline without any contrastive pretraining.
Overall, SPADE consistently outperformed both the baseline and single-expert variant across most cancer types, achieving the highest AUC and F1 scores in six out of eight datasets. The improvement over the Virchow2 baseline highlights that the mixture-of-experts design and contrastive alignment provide complementary benefits even when using other robust histology foundation models as the image encoder backbone.
Although absolute performance was slightly lower than with the UNI backbone (Table \ref{table:subt_res}), the trends remain consistent, indicating that SPADE’s architecture and training strategy are robust to the choice of foundation model. These findings suggest that SPADE can effectively adapt to diverse histology feature encoders.

\begin{table}[H]
\centering
\resizebox{\textwidth}{!}{
\begin{tabular}{l|cccc|cccc|cc|cc|cccc}
                        & \multicolumn{4}{c|}{Lung}                                                       & \multicolumn{4}{c|}{Prostate}                                                                                                  & \multicolumn{2}{c|}{\begin{tabular}[c]{@{}c@{}}Lymph\\ Nodes\end{tabular}} & \multicolumn{2}{c|}{Ovarian}                                              & \multicolumn{4}{c}{Breast}                                                                                                    \\ \hline
                        & \multicolumn{2}{c|}{PLCO}                          & \multicolumn{2}{c|}{CPTAC} & \multicolumn{2}{c|}{PANDA}                         & \multicolumn{2}{c|}{\begin{tabular}[c]{@{}c@{}}TCGA-\\ PRAD\end{tabular}} & \multicolumn{2}{c|}{CAM16}                                                 & \multicolumn{2}{c|}{\begin{tabular}[c]{@{}c@{}}UBC-\\ OCEAN\end{tabular}} & \multicolumn{2}{c|}{PLCO}                          & \multicolumn{2}{c}{\begin{tabular}[c]{@{}c@{}}TCGA-\\ BRCA\end{tabular}} \\ \hline
                        & AUC           & \multicolumn{1}{c|}{F1}            & AUC           & F1         & AUC           & \multicolumn{1}{c|}{F1}            & AUC                                 & F1                                  & AUC                                  & F1                                  & AUC                                 & F1                                  & AUC           & \multicolumn{1}{c|}{F1}            & AUC                                 & F1                                 \\ \hline
Virchow2+ABMIL          & 93.1          & \multicolumn{1}{c|}{80.0}          & 96.1          & 90.0       & {\ul 93.6}    & \multicolumn{1}{c|}{{\ul 68.6}}    & 81.8                                & 41.1                                & {\ul 83.7}                           & {\ul 57.9}                          & {\ul 97.9}                          & {\ul 89.7}                          & 76.4          & \multicolumn{1}{c|}{58.9}          & 97.5                                & {\ul 96.8}                         \\
SPADE(Virchow2)-0+ABMIL & {\ul 96.4}    & \multicolumn{1}{c|}{{\ul 88.4}}    & 95.5          & 87.8       & 92.9          & \multicolumn{1}{c|}{66.5}          & {\ul 81.9}                          & \textbf{55.3}                       & 70.1                                 & {\ul 57.9}                          & 97.6                                & 89.4                                & {\ul 79.1}    & \multicolumn{1}{c|}{{\ul 61.4}}    & \textbf{98.3}                       & {\ul 96.8}                         \\
SPADE(Virchow2)+ABMIL   & \textbf{98.1} & \multicolumn{1}{c|}{\textbf{93.0}} & \textbf{98.9} & {\ul 96.3} & \textbf{94.1} & \multicolumn{1}{c|}{\textbf{69.0}} & \textbf{82.5}                       & {\ul 48.2}                          & \textbf{95.8}                        & \textbf{78.7}                       & \textbf{98.4}                       & \textbf{93.0}                       & \textbf{79.6} & \multicolumn{1}{c|}{\textbf{63.1}} & {\ul 98.2}                          & \textbf{97.8}                      \\ \hline
\end{tabular}
}
\caption{\textbf{Cancer suptyping results with Virchow2.} 
Comparison of the mixture-of-experts model (SPADE) to the single-expert variant (SPADE-0), both using Virchow foundation model–derived histology features instead of UNI. The comparison also includes the Virchow2 baseline without any contrastive pretraining. Results are reported in terms of AUC and F1 score (\%). The best performance is shown in bold, and the second best is underlined.}
\label{tab:virch_subt}
\end{table}

 \bibliographystyle{elsarticle-num} 
 \bibliography{bib.bib}

\begin{thebibliography}{10}
\expandafter\ifx\csname url\endcsname\relax
  \def\url#1{\texttt{#1}}\fi
\expandafter\ifx\csname urlprefix\endcsname\relax\def\urlprefix{URL }\fi
\expandafter\ifx\csname href\endcsname\relax
  \def\href#1#2{#2} \def\path#1{#1}\fi

\bibitem{wang2022cell}
Y.~Wang, Y.~G. Wang, C.~Hu, M.~Li, Y.~Fan, N.~Otter, I.~Sam, H.~Gou, Y.~Hu, T.~Kwok, et~al., Cell graph neural networks enable the precise prediction of patient survival in gastric cancer, NPJ precision oncology 6~(1) (2022) 45.

\bibitem{li2021multi}
J.~Li, W.~Li, A.~Sisk, H.~Ye, W.~D. Wallace, W.~Speier, C.~W. Arnold, A multi-resolution model for histopathology image classification and localization with multiple instance learning, Computers in biology and medicine 131 (2021) 104253.

\bibitem{saednia2022quantitative}
K.~Saednia, A.~Lagree, M.~A. Alera, L.~Fleshner, A.~Shiner, E.~Law, B.~Law, D.~W. Dodington, F.-I. Lu, W.~T. Tran, et~al., Quantitative digital histopathology and machine learning to predict pathological complete response to chemotherapy in breast cancer patients using pre-treatment tumor biopsies, Scientific Reports 12~(1) (2022) 9690.

\bibitem{howard2023integration}
F.~M. Howard, J.~Dolezal, S.~Kochanny, G.~Khramtsova, J.~Vickery, A.~Srisuwananukorn, A.~Woodard, N.~Chen, R.~Nanda, C.~M. Perou, et~al., Integration of clinical features and deep learning on pathology for the prediction of breast cancer recurrence assays and risk of recurrence, NPJ Breast Cancer 9~(1) (2023) 25.

\bibitem{wang2024pathology}
X.~Wang, J.~Zhao, E.~Marostica, W.~Yuan, J.~Jin, J.~Zhang, R.~Li, H.~Tang, K.~Wang, Y.~Li, et~al., A pathology foundation model for cancer diagnosis and prognosis prediction, Nature (2024) 1--9.

\bibitem{chen2024towards}
R.~J. Chen, T.~Ding, M.~Y. Lu, D.~F. Williamson, G.~Jaume, A.~H. Song, B.~Chen, A.~Zhang, D.~Shao, M.~Shaban, et~al., Towards a general-purpose foundation model for computational pathology, Nature Medicine 30~(3) (2024) 850--862.

\bibitem{chen2022scaling}
R.~J. Chen, C.~Chen, Y.~Li, T.~Y. Chen, A.~D. Trister, R.~G. Krishnan, F.~Mahmood, Scaling vision transformers to gigapixel images via hierarchical self-supervised learning, in: Proceedings of the IEEE/CVF Conference on Computer Vision and Pattern Recognition, 2022, pp. 16144--16155.

\bibitem{xu2024whole}
H.~Xu, N.~Usuyama, J.~Bagga, S.~Zhang, R.~Rao, T.~Naumann, C.~Wong, Z.~Gero, J.~Gonz{\'a}lez, Y.~Gu, et~al., A whole-slide foundation model for digital pathology from real-world data, Nature (2024) 1--8.

\bibitem{moffitt2022emerging}
J.~R. Moffitt, E.~Lundberg, H.~Heyn, The emerging landscape of spatial profiling technologies, Nature Reviews Genetics 23~(12) (2022) 741--759.

\bibitem{moses2022museum}
L.~Moses, L.~Pachter, Museum of spatial transcriptomics, Nature methods 19~(5) (2022) 534--546.

\bibitem{lee2023promise}
R.~Y. Lee, C.~W. Ng, M.~P. Rajapakse, N.~Ang, J.~P.~S. Yeong, M.~C. Lau, The promise and challenge of spatial omics in dissecting tumour microenvironment and the role of ai, Frontiers in Oncology 13 (2023) 1172314.

\bibitem{rao2021exploring}
A.~Rao, D.~Barkley, G.~S. Fran{\c{c}}a, I.~Yanai, Exploring tissue architecture using spatial transcriptomics, Nature 596~(7871) (2021) 211--220.

\bibitem{xun2023reconstruction}
Z.~Xun, X.~Ding, Y.~Zhang, B.~Zhang, S.~Lai, D.~Zou, J.~Zheng, G.~Chen, B.~Su, L.~Han, et~al., Reconstruction of the tumor spatial microenvironment along the malignant-boundary-nonmalignant axis, Nature Communications 14~(1) (2023) 933.

\bibitem{jaume2024hest}
G.~Jaume, P.~Doucet, A.~H. Song, M.~Y. Lu, C.~Almagro-P{\'e}rez, S.~J. Wagner, A.~J. Vaidya, R.~J. Chen, D.~F. Williamson, A.~Kim, et~al., Hest-1k: A dataset for spatial transcriptomics and histology image analysis, arXiv preprint arXiv:2406.16192 (2024).

\bibitem{oord2018representation}
A.~v.~d. Oord, Y.~Li, O.~Vinyals, Representation learning with contrastive predictive coding, arXiv preprint arXiv:1807.03748 (2018).

\bibitem{radford2021learning}
A.~Radford, J.~W. Kim, C.~Hallacy, A.~Ramesh, G.~Goh, S.~Agarwal, G.~Sastry, A.~Askell, P.~Mishkin, J.~Clark, et~al., Learning transferable visual models from natural language supervision, in: International conference on machine learning, PMLR, 2021, pp. 8748--8763.

\bibitem{xie2024spatially}
R.~Xie, K.~Pang, S.~Chung, C.~Perciani, S.~MacParland, B.~Wang, G.~Bader, Spatially resolved gene expression prediction from histology images via bi-modal contrastive learning, Advances in Neural Information Processing Systems 36 (2024).

\bibitem{jaume2024transcriptomics}
G.~Jaume, L.~Oldenburg, A.~Vaidya, R.~J. Chen, D.~F. Williamson, T.~Peeters, A.~H. Song, F.~Mahmood, Transcriptomics-guided slide representation learning in computational pathology, in: Proceedings of the IEEE/CVF Conference on Computer Vision and Pattern Recognition, 2024, pp. 9632--9644.

\bibitem{ma2024mode}
J.~Ma, P.-Y. Huang, S.~Xie, S.-W. Li, L.~Zettlemoyer, S.-F. Chang, W.-T. Yih, H.~Xu, Mode: Clip data experts via clustering, in: Proceedings of the IEEE/CVF Conference on Computer Vision and Pattern Recognition, 2024, pp. 26354--26363.

\bibitem{vo2024automatic}
H.~V. Vo, V.~Khalidov, T.~Darcet, T.~Moutakanni, N.~Smetanin, M.~Szafraniec, H.~Touvron, C.~Couprie, M.~Oquab, A.~Joulin, et~al., Automatic data curation for self-supervised learning: A clustering-based approach, arXiv preprint arXiv:2405.15613 (2024).

\bibitem{ilse2018attention}
M.~Ilse, J.~Tomczak, M.~Welling, Attention-based deep multiple instance learning, in: International conference on machine learning, PMLR, 2018, pp. 2127--2136.

\bibitem{wolf2018scanpy}
F.~A. Wolf, P.~Angerer, F.~J. Theis, Scanpy: large-scale single-cell gene expression data analysis, Genome biology 19 (2018) 1--5.

\bibitem{zhu2013prostate}
C.~S. Zhu, P.~F. Pinsky, B.~S. Kramer, P.~C. Prorok, M.~P. Purdue, C.~D. Berg, J.~K. Gohagan, The prostate, lung, colorectal, and ovarian cancer screening trial and its associated research resource, Journal of the National Cancer Institute 105~(22) (2013) 1684--1693.

\bibitem{grossman2016toward}
R.~L. Grossman, A.~P. Heath, V.~Ferretti, H.~E. Varmus, D.~R. Lowy, W.~A. Kibbe, L.~M. Staudt, Toward a shared vision for cancer genomic data, New England Journal of Medicine 375~(12) (2016) 1109--1112.

\bibitem{bulten2022artificial}
W.~Bulten, K.~Kartasalo, P.-H.~C. Chen, P.~Str{\"o}m, H.~Pinckaers, K.~Nagpal, Y.~Cai, D.~F. Steiner, H.~Van~Boven, R.~Vink, et~al., Artificial intelligence for diagnosis and gleason grading of prostate cancer: the panda challenge, Nature medicine 28~(1) (2022) 154--163.

\bibitem{bejnordi2017diagnostic}
B.~E. Bejnordi, M.~Veta, P.~J. Van~Diest, B.~Van~Ginneken, N.~Karssemeijer, G.~Litjens, J.~A. Van Der~Laak, M.~Hermsen, Q.~F. Manson, M.~Balkenhol, et~al., Diagnostic assessment of deep learning algorithms for detection of lymph node metastases in women with breast cancer, Jama 318~(22) (2017) 2199--2210.

\bibitem{litjens20181399}
G.~Litjens, P.~Bandi, B.~Ehteshami~Bejnordi, O.~Geessink, M.~Balkenhol, P.~Bult, A.~Halilovic, M.~Hermsen, R.~Van~de Loo, R.~Vogels, et~al., 1399 h\&e-stained sentinel lymph node sections of breast cancer patients: the camelyon dataset, GigaScience 7~(6) (2018) giy065.

\bibitem{farahani2022deep}
H.~Farahani, J.~Boschman, D.~Farnell, A.~Darbandsari, A.~Zhang, P.~Ahmadvand, S.~J. Jones, D.~Huntsman, M.~K{\"o}bel, C.~B. Gilks, et~al., Deep learning-based histotype diagnosis of ovarian carcinoma whole-slide pathology images, Modern Pathology 35~(12) (2022) 1983--1990.

\bibitem{asadi2024machine}
M.~Asadi-Aghbolaghi, H.~Farahani, A.~Zhang, A.~Akbari, S.~Kim, A.~Chow, S.~Dane, O.~C. Consortium, O.~Consortium, D.~G~Huntsman, et~al., Machine learning-driven histotype diagnosis of ovarian carcinoma: Insights from the ocean ai challenge, medRxiv (2024) 2024--04.

\bibitem{kather2020pan}
J.~N. Kather, L.~R. Heij, H.~I. Grabsch, C.~Loeffler, A.~Echle, H.~S. Muti, J.~Krause, J.~M. Niehues, K.~A. Sommer, P.~Bankhead, et~al., Pan-cancer image-based detection of clinically actionable genetic alterations, Nature cancer 1~(8) (2020) 789--799.

\bibitem{lu2024visual}
M.~Y. Lu, B.~Chen, D.~F. Williamson, R.~J. Chen, I.~Liang, T.~Ding, G.~Jaume, I.~Odintsov, L.~P. Le, G.~Gerber, et~al., A visual-language foundation model for computational pathology, Nature Medicine 30~(3) (2024) 863--874.

\bibitem{zimmermann2024virchow2}
E.~Zimmermann, E.~Vorontsov, J.~Viret, A.~Casson, M.~Zelechowski, G.~Shaikovski, N.~Tenenholtz, J.~Hall, D.~Klimstra, R.~Yousfi, et~al., Virchow2: Scaling self-supervised mixed magnification models in pathology, arXiv preprint arXiv:2408.00738 (2024).

\bibitem{liu2018integrated}
J.~Liu, T.~Lichtenberg, K.~A. Hoadley, L.~M. Poisson, A.~J. Lazar, A.~D. Cherniack, A.~J. Kovatich, C.~C. Benz, D.~A. Levine, A.~V. Lee, et~al., An integrated tcga pan-cancer clinical data resource to drive high-quality survival outcome analytics, Cell 173~(2) (2018) 400--416.

\bibitem{epstein20162014}
J.~I. Epstein, L.~Egevad, M.~B. Amin, B.~Delahunt, J.~R. Srigley, P.~A. Humphrey, G.~Committee, et~al., The 2014 international society of urological pathology (isup) consensus conference on gleason grading of prostatic carcinoma: definition of grading patterns and proposal for a new grading system, The American journal of surgical pathology 40~(2) (2016) 244--252.

\bibitem{suurmeijer2018novel}
A.~J. Suurmeijer, B.~C. Dickson, D.~Swanson, L.~Zhang, Y.-S. Sung, P.~Cotzia, C.~D. Fletcher, C.~R. Antonescu, A novel group of spindle cell tumors defined by s100 and cd34 co-expression shows recurrent fusions involving raf1, braf, and ntrk1/2 genes, Genes, Chromosomes and Cancer 57~(12) (2018) 611--621.

\bibitem{cao2024stformer}
S.~Cao, K.~Yang, J.~Cheng, J.~Li, H.-B. Shen, X.~Pan, Y.~Yuan, stformer: a foundation model for spatial transcriptomics, bioRxiv (2024) 2024--09.

\bibitem{zhao2025stofm}
S.~Zhao, Y.~Luo, G.~Yang, Y.~Zhong, H.~Zhou, Z.~Nie, Stofm: a multi-scale foundation model for spatial transcriptomics, arXiv preprint arXiv:2507.11588 (2025).

\end{thebibliography}






\end{document}